\documentclass[nojss]{jss}

\usepackage{thumbpdf,lmodern}

\usepackage{amsmath,amsfonts,amssymb,amscd,xspace}
\usepackage{booktabs}
\usepackage{multirow}
\usepackage[english]{babel}
\usepackage[utf8]{inputenc}

\usepackage{framed}

\usepackage{setspace}

\author{Berent Ånund Strømnes Lunde\\University of Stavanger
   \And Tore Selland Kleppe\\University of Stavanger
}
\Plainauthor{Berent Lunde, Tore Kleppe}

\title{\pkg{agtboost}: Adaptive and Automatic Gradient Tree Boosting Computations}
\Plaintitle{agtboost: Adaptive and Automatic Gradient Boosting Computations}
\Shorttitle{\pkg{agtboost}: Automatic Function Estimation}

\Abstract{
\pkg{agtboost} is an \proglang{R} package implementing fast gradient 
tree boosting computations in a manner similar to other established frameworks such 
as \pkg{xgboost} and \pkg{LightGBM}, but with significant decreases in computation time and required mathematical and technical knowledge. 
The package automatically takes care of split/no-split decisions and selects the number of trees
in the gradient tree boosting ensemble, i.e., \pkg{agtboost} adapts the complexity of the ensemble automatically
to the information in the data.
All of this is done during a single training run, which is made possible 
by utilizing developments in information theory for tree algorithms \citep{lunde2020information}.
\pkg{agtboost} also comes with a feature importance function that eliminates the common practice of inserting noise features.
Further, a useful model validation function performs the Kolmogorov-Smirnov test on the learned distribution.
}

\Plainkeywords{gradient tree boosting, information criterion, automatic function estimation, R}

\Address{
  Berent Ånund Strømnes Lunde\\
  Department of Mathematics and Physics\\
  Faculty of Science and Technology\\
  University of Stavanger\\
  Kristine Bonnevies vei 22\\
  4021 Stavanger, Norway\\
  E-mail: \email{berent.a.lunde@uis.no}\\
}

\begin{document}



\section[Introduction]{Introduction: Tuning of gradient tree boosting} \label{sec:intro}
Gradient tree boosting (GTB) \citep{friedman2001greedy, mason1999boosting} has risen to prominence for regression problems after the introduction of \pkg{xgboost} \citep{chen2016xgboost}.
The GTB model is an ensemble-type model, that consist of classification and regression trees (CART) \citep{breiman1984classification} that are learned in an iterative manner.
GTB models are very flexible in that they automatically learn non-linear relationships and interaction effects.
However, with the increased flexibility of GTB models comes substantial worries of overfitting.
The top performing gradient tree boosting libraries, such as \pkg{xgboost}, \pkg{LightGBM} \citep{ke2017lightgbm} and \pkg{catboost} \citep{dorogush2018catboost}, 
all come with a large number of hyperparameters available for manual tuning
to constrain the complexity of the GTB models.
Training of gradient tree boosting models, in general, thus require some familiarity with both
the chosen package, and the data for efficient tuning and application to the problem at hand. 

The main focus of the hyperparameters and tuning are to solve the following problems:
\begin{itemize}
	\item \textbf{The complexity of trees:} What are the topology of all the different trees? Too complex trees overfits, while simple stump-models cannot capture interaction effects. This is typically solved using a hyperparameter that penalizes (equally) the number of leaves in the tree. \pkg{xgboost} hyperparameters for this are \code{gamma}, \code{max\_depth}, \code{min\_child\_weight}, and \code{max\_leaves}.
	\item \textbf{The number of trees:} How many iterations should the tree-boosting algorithm do before terminating? Too early stopping will leave information unlearned, while too late stopping will see the last trees adapting mostly to noise. An early-stopping hyperparameter is usually tuned to obtain ensembles of adequate size. Tuned in \pkg{xgboost} with \code{nrounds}.
	\item \textbf{Making space for feature trees to learn:}
	If each tree is optimized alone, early trees will have a tendency to learn additive relationships and information that subsequent trees could learn more efficiently \citep{friedman2000additive}.
	An additional downside of large early trees is difficult model-interpretability.
	The hyperparameter solution typically involves tuning the maximum depth (\code{max\_depth} in \pkg{xgboost}) globally for all trees.
\end{itemize}
The five parameters of \pkg{xgboost} mentioned above are typically selected as the top-performing parameters found from $k$-fold cross validation (CV) \citep{stone1974cross}.
CV, however, increases computation times extensively, and requires more work through coding and knowledge on the part of the user.

\pkg{agtboost} is an implementation of the theory in \citet{lunde2020information}, which unlocks 
computationally fast and automatic solutions to the problems listed above, and as a consequence removes selection of hyperparameters through CV from the problem. The key is an information criterion that can be applied after the greedy binary-splitting profiling procedure used in learning trees. 
The theory is built upon maximal selection of chi-squared statistics \citep{white1982maximum, gombay1990asymptotic} and the convergence of an empirical process to a continuous time stochastic process.
\citet{lunde2020information} subsequently discuss how both tree-size and the number of trees then can be chosen automatically.
This paper supplements \citet{lunde2020information} by describing a package built on this theory, i.e., \pkg{agtboost}.
Some new innovations are also introduced, which all have their basis in the information criterion. In general, they address the problems with feature importance and the optimization of trees alone.

Note that there exist other hyperparameters that may increase accuracy (but that are not vital) for GTB models. Most notably are parameters for a regularized objective (see e.g., \citet{chen2016xgboost}) and stochastic sampling of observations 
during boosting iterations (for an overview, see \citet{friedman2001elements} and for recent innovations for GTB see \citet{ke2017lightgbm}). These features are not yet implemented, but subject to further
research, as more work is required to adhere to the philosophy of \pkg{agtboost} --
that all hyperparameters should be automatically tuned. 

This paper starts by introducing gradient tree boosting in Section \ref{sec:gtboosting} and the information criterion in Section \ref{sec:information}, and proceeds with the innovations and software implementation in Section \ref{sec:software}. Section \ref{sec:using} describes \pkg{agtboost} from a user's perspective. Section \ref{sec:case} studies and compares the different variants of
\pkg{agtboost} models for the large sized Higgs dataset.
Finally, Section \ref{sec:discussion} discusses and concludes.



\section{Gradient tree boosting} \label{sec:gtboosting}
This and the following section closely follow the setup of \citet{lunde2020information}.
The (typical) objective of gradient tree boosting procedures is the supervised
learning problem
\begin{align}\label{eq:supervised-objective}
	f(x) = \arg\min_f E[l(y,f(x))],
\end{align}
for a response $y\in R$, feature vector $x\in R^m$, and loss function $l$ measuring the difference between the response and prediction $\hat y = f(x)$.
For gradient boosting to work, we require $l$ to be both differentiable and convex.
Then, using training data, say $\mathcal{D}_{n}=\{(y_i,x_i)\}_{i=1}^n$, we seek to approximate (\ref{eq:supervised-objective}) 
by learning $f$ in an iterative manner: Given a function $f^{(k-1)}$, we seek $f_k$
to minimize
\begin{align}\label{eq:supervised-objective-iterative}
f_k(x) = \arg\min_{f_k} E[l(y,f^{(k-1)}(x)+f_k(x))],
\end{align}
approximately to second order. This is done by computing the derivatives
\begin{align}
	g_{i,k} = \left.\frac{\partial}{\partial \hat{y}^{(k-1)}}l(y,\hat{y}^{(k-1)})\right|_{\hat{y}^{(k-1)} = f^{(k-1)}(x)}
	,~
	h_{i,k} = \left.\frac{\partial^2}{\partial (\hat{y}^{(k-1)})^2}l(y,\hat{y}^{(k-1)})\right|_{\hat{y}^{(k-1)} = f^{(k-1)}(x)},
\end{align}
for each observation in the training data, and then approximate the expected loss by averaging,
\begin{align}\label{eq:gradient-boosting-objective}
	f_k(x) = \arg\min_f \frac{1}{n}\sum_{i=1}^{n} l(y,\hat{y}_i^{(k-1)})
	+ g_{i,k}f_k(x) + \frac{1}{2}h_{i,k}f_k(x)^2.
\end{align}
Terminating this procedure at iteration $K$, the final model is an additive model of the form
\begin{align}
	\hat{y}_i = \hat{y}^{(K)} = \sum_{k=1}^{K} f_k(x_i).
\end{align}

Still, \eqref{eq:gradient-boosting-objective} is a hard problem, as the search among all possible functions is obviously infeasible. Therefore, it is necessary to constrain the search over a a family of functions, or "base learners". 
While multiple choices exist, \pkg{agtboost} follows the convention of using CART \citep{breiman1984classification}.

For a full discussion of decision trees, see \citep{friedman2001elements} for a general treatment, and \citet{lunde2020information} for details on its use in gradient tree boosting and \pkg{agtboost}.
We constrain ourselves to a brief mention of important aspects.
Firstly, decision trees learn constant predictions (called leaf-weights) in regions of
feature space. 
We let $I_{t,k}=\{i:q_k(x)=t\}$ denote the index-set of training indices that falls into region (or leaf) $t$, denoted by $q_k(x)=t$, where $q_k$ is the topology of the $k$'th tree, a function that takes the feature vector and returns the the node-index or corresponding index of region in feature space, $t$.
The prediction from the tree is given by 
\begin{align}
	f_k(x_i) = w_{q_k(x_i)}.
\end{align}
Secondly, the estimated leaf-weights have closed form in the 2'nd order boosting
procedure described above, namely
\begin{align}\label{eq:leaf-weights}
	\hat{w}_{t,k} = -\frac{G_{t,k}}{H_{t,k}}
	\:\text{where}\:
	G_{t,k} = \sum_{i\in I_{t,k}}g_{i,k}
	\:\text{and}\:
	H_{t,k} = \sum_{i\in I_{t,k}} h_{i,k}.
\end{align}
Thirdly, the regions of feature space are learned by iteratively splitting all leaf-regions (starting with the full feature space as the only leaf) creating new leaves and regions. The region is split on the split-point that gives the largest reduction in training loss, 
say $R_t$, among all possible binary splits. This search is fast, as the training loss, 
modulus unimportant constant terms, in region $t$ is given as 
\begin{align}
	l_t = -\frac{G_{t,k}^2}{2nH_{t,k}},
\end{align}
and enumeration of possible splits can therefore be done in $mn\log n$ time.

The above procedure creates the decision tree $f_k$, which is then added to the model $f^{(k-1)}$ by
\begin{align}
	f^{(k)} = f^{(k-1)} + \delta f_k,~
	\delta \in (0,1).
\end{align}
The constant $\delta$, typically called the "learning rate" or "shrinkage", leaves space for feature models to learn. Values are often taken as "small", but this comes at the added computational cost of an increase in the number of boosting iterations (infinite when $\delta\to 0$) before convergence.
The learning rate is the only hyperparameter of the boosting procedure in \pkg{agtboost}
that is not tuned automatically.
The default value is set at $0.01$, which should be sufficiently small for most applications 
without incurring too much computational cost.

\section{Information criteria} \label{sec:information}
This section introduces generalization loss-based information criteria, which includes
types such as Akaike Information Criterion (AIC) \citep{akaike1974new}, Takeuchi Information Criterion (TIC) \citep{takeuchi1976distribution} and Network Information Criterion (NIC) \citep{murata1994network}.
Generalization loss is perhaps better known as an instance of test loss, say $l(y^0,f(x^0))$, where 
$(y^0,x^0)$ is an observation unseen in the training phase, i.e., not an element and independent of $\mathcal{D}_n$.
This specification is important, as (\ref{eq:supervised-objective}) is intended for 
this quantity, and using the training loss, in our case (\ref{eq:gradient-boosting-objective}) as an estimator, care must be taken as the training loss is biased downwards in expectation.
This bias is known as the \textit{optimism} of the training loss \citep{friedman2001elements},
and denoted $C$.
A generalization loss-based information criterion, say $\tilde{C}$, is intended 
to capture the size of the optimism, such that adding $\tilde{C}$ to the training loss gives an (at least approximately) unbiased estimator of the expected generalization loss. 

The equation behind the generalization loss-based information criteria mentioned above is
\begin{align}\label{eq:information-criteria-equation}
	\tilde{C} &= 
	\texttt{tr}\left(E\left[\nabla_{\theta}^2l\left(y,f(x;\theta_0)\right)\right]Cov(\hat{\theta})\right),\\
	\tilde{C} &\approx E[l(y^0,f(x^0;\hat{\theta}))]
	-	E[l(y_1,f(x_1;\hat{\theta}))],\notag
\end{align}
where $(y_1,x_1)$ is a (random) instance of the training dataset, and $\theta_0$ is the population minimizer of (\ref{eq:supervised-objective}) where optimization is done over a parametric family of functions and $\hat{\theta}$ is not at the boundary of parameter space \citep[Eqn. 7.32]{burnham2003model}.

In the GTB training procedure described in section \ref{sec:gtboosting}, the model selection questions
of added complexity are always at the "local" root (constant prediction) versus stump (a tree with two leaf-nodes) model.
It is always one, and only one, node at a time that is tried split in this gradient tree boosting procedure.
Thus isolate this relevant node by assuming and conditioning on that $q(x)=t$, then node $t$ may be referenced and treated as a root.
Further, for convenient notation, assume to be at some boosting iteration $k$, thus dropping notational dependence on current iteration, and
let $l$ and $r$ denote the left and right child nodes of node $t$.
The idea of \citet{lunde2020information} is to use the optimism of the root (node $t$, conditioned on data falling into leaf $t$, i.e., $q(x)=t$) and stump model (split of node $t$, with the same conditioning as for the root), say 
$C_\text{root}$ and $C_\text{stump}$, to adjust the root and stump training loss, to see if there is expected a positive reduction in generalization loss,
\begin{align}
	-\frac{1}{2n}\left[
	\frac{G_t^2}{H_t} 
	- \left(\frac{G_{l}^2}{H_{l}} + \frac{G_{r}^2}{H_{r}} \right)
	\right]
	+ C_\text{root} - C_\text{stump} > 0,
\end{align}
which would be used to decide if to split further in a tree, and to see if a tree-stump model $f_k$ could be added to the ensemble $f^{k-1}$.
The root and stump training loss combined in the parenthesis is the loss-reduction, $R$,
that is profiled over for different binary splits.
This profiling, however, complicates evaluation of the above inequality as it induces optimism
into $C_{stump}$ which \eqref{eq:information-criteria-equation} cannot handle directly.

We may combine the root and stump optimisms to create
a loss-reduction optimism, say $C_R=C_\text{root} - C_\text{stump}$.
\citet{lunde2020information} derives the following estimator for the optimism of loss-reduction
\begin{align}\label{eq:information-criteria-loss-reduction}
\tilde{C}_R = -\tilde{C}_{root}\pi_tE\left[ B(\mathbf{x})  \right],
\end{align}
where $\pi_t$ is the probability of being in node $t$, estimated by the fraction of training data passed to node $t$, and $\tilde{C}_{root}$ is calculated as \eqref{eq:information-criteria-equation} conditioned on being in node $t$, and may be estimated using the sandwich estimator \citep{huber1967behavior} together with the empirical Hessian.
The remaining quantity in \eqref{eq:information-criteria-loss-reduction}, $B(\mathbf{x})$ where $\mathbf{x}$ is the full design matrix of the training data,
is a maximally chosen random variable
\begin{align}
B(\mathbf{x}) = \max_{j\leq 1\leq m} B_j(x_{\cdot j}),~
B_j(x_{\cdot j}) = \max_{1\leq k \leq a} S_j(\tau_k)
\end{align}
defined from a specification of the Cox-Ingersoll-Ross process (CIR) \citep{cox1985theory}, $S(\tau)$, with speed of adjustment to the mean $2$, long term mean $1$ and instantaneous rate of volatility $2\sqrt{2}$, therefore having dynamics given by the SDE
\begin{align}
dS(\tau) = 2(1-S(\tau))dt + 2\sqrt{2S(\tau)}dW(\tau),
\end{align}
which is "observed" at time-points $\tau_k = \frac{1}{2}\log\frac{u_k(1-\epsilon)}{\epsilon(1-u_k)},\: \epsilon\to0$ defined from the split-points $u_k = p(x_{ij}\leq s_k)$, $i=1:(n-1)$ on feature $j$.
To estimate $E[B(\mathbf{x})]$ is, however, not straight forward.
\citet{lunde2020information} discuss a solution using exact simulation of the CIR process using $\epsilon=10^{-7}$, together with the fact that the stationary CIR is in the maximum domain of attraction of the Gumbel distribution \citep{gombay1990asymptotic}, an independence assumption on features, and numerical integration to evaluate the expectation.

\section{Software implementation and innovations} \label{sec:software}
\pkg{agtboost} employs the r.h.s. of (\ref{eq:information-criteria-loss-reduction}) to solve the list of
problems presented in the introduction, which for other implementations are solved by 
tuning the hyperparameters using techniques such as cross-validation (CV) \citep{stone1974cross}.
The use of (\ref{eq:information-criteria-loss-reduction}) alleviates the need for hyperparameters tuned
with CV, as it allows the base-learner trees $f_k$ and the ensemble $f^{(k)}$ to stop at a given complexity
that is adapted to the training data at hand.
Thus significantly increasing the speed of training a gradient tree boosting ensemble.
Furthermore, the technical knowledge imposed on the user, with respect to both gradient tree boosting and the dataset at hand, is reduced. \pkg{agtboost} is coded in \proglang{C++} for fast computations, and relies on \pkg{Eigen} \citep{eigenweb} for linear algebra,
the \proglang{R} header files \citep{Rlanguage2018} for some distributions, and \pkg{Rcpp} \citep{rcpp2011R} for bindings to \proglang{R}.
The remainder of this section goes through the innovations in \pkg{agtboost} that 
directly attacks the previously mentioned tuning-problems.

\subsection{Adaptive tree size} \label{subsec:adaptive}
\begin{figure}[t!]
	\centering
	\includegraphics[width=1.0\textwidth,height=7cm]{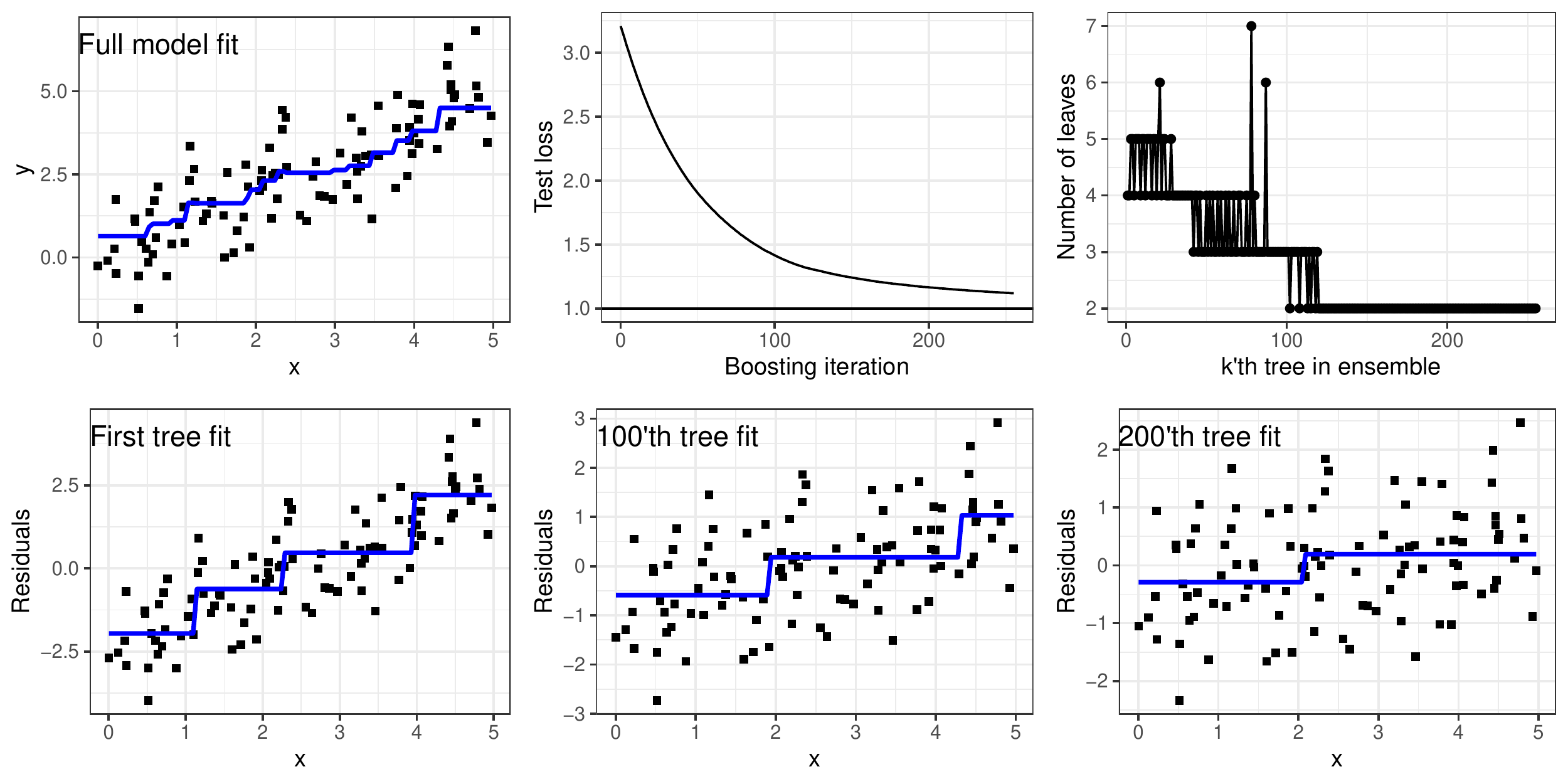}
	\caption{\label{fig:tree_model_fits} 
		Top: [left] Full model fit on training-data, [middle] test loss versus the number of boosting iterations, with the minimum expected loss possible (true model) included with a horizontal line, and [right] the number of leaves in the $k$'th tree.
		Bottom: The First, 100'th and 200'th tree fits to the training MSE-residuals, $r_i^{(k)}=-g_{i,k}/h_{i,k}=y_i-\hat{y}^{(k-1)}$ at respective iterations.
		The $n=100$ training and test observations were i.i.d. generated with a linear structural relationship and Gaussian noise, $x\sim U(0,5)$ and $y\sim N(x,1)$. 
		}
\end{figure}
Equipped with an information criterion for loss reduction after greedy-split-profiling,
the necessary adjustments are rather straight-forward, and are also discussed in \citet{lunde2020information}.
For completeness, the usage of \eqref{eq:information-criteria-loss-reduction} towards selecting the complexity of trees $f_k$ is restated here:
After the split that maximizes training loss reduction $R$ is found, the following inequality is tested
\begin{align}\label{eq:adaptive-tree-inequality}
	R + \tilde{C}_{R} > 0,
\end{align}
and if it evaluates to \code{TRUE}, then two new leaves (and regions) are created
and successive splitting on them is performed. This continues until (\ref{eq:adaptive-tree-inequality}) evaluates to \code{FALSE}. This criterion is employed for all but the first (root) split, which is 
forced. This forced split is done to assure some increase in model complexity, as $\sum_i g_i=0$ is always true
and a root model will therefore be equivalent to adding zero to the model.

Figure \ref{fig:tree_model_fits} illustrates this adaptivity: Visually, it is seen that trees assign higher leaf-node predictions to observations with high values of $x$ than to small values of $x$, thus capturing the structural relationship $y=x$ in the simulated data.
Furthermore, of the trees plotted in the lower lane together with residuals 
$y_i-\hat{y}_i^{(k-1)}=-g_{i,k}/h_{i,k}$ 
at iterations $k\in\{ 1, 100, 200 \}$, none of them can be seen to be complex enough
to adapt to the Gaussian noise in the training data.
Indeed, every subsequent iteration reduces the value of test loss, seen from the top-middle panel.
Further verification of this is given by the decreasing complexity of trees (measured in terms of number of leaves), corresponding to residuals at early iterations necessarily containing more information than later ones.
Thus, early trees therefore tend to be more complex than at later stages of the training.

\subsection{Automatic early stopping} \label{subsec:automatic}
The natural stopping criterion for the iterative boosting procedure, in the context 
of supervised learning, is to stop when
the increase in model complexity no longer gives a reduction in generalization loss. 
From Figure \ref{fig:tree_model_fits}, the top-right plot shows that the later
iterations tend to be tree-stumps.
Indeed, a tree constructed using the method described in Section \ref{subsec:adaptive}
will be a tree stump at the iteration where the natural stopping criterion terminates
the algorithm.
This is because a more complex tree must have
passed the "barrier" (i.e., inequality) \eqref{eq:adaptive-tree-inequality},
and necessarily will have a decrease in generalization loss, as long
as $\delta\in(0,1]$.
Care must however be taken, as we are scaling the $k$'th tree with the learning rate, $\delta$,
and \ref{eq:adaptive-tree-inequality} might therefore not be used directly.
\citet{lunde2020information} discuss this: The solution lies in the general equation for Equation
\eqref{eq:information-criteria-equation} \citep{friedman2001elements}
\begin{align}
	\tilde{C} = \frac{2}{n}\sum_{i=1}^{n}Cov(y_i,\hat y_i),
\end{align}
and the optimism therefore scales linearly. The training loss, on the other hand, does not
and can be seen to scale with the factor $\delta(2-\delta)$ from direct computations.
Scaling the training loss and optimism appropriately, \pkg{agtboost} evaluates a similar inequality as (\ref{eq:adaptive-tree-inequality}), namely
\begin{align}\label{eq:automatic-early-stopping}
	\delta(2-\delta)R + \delta \tilde{C}_{R} > 0,
\end{align}
which if evaluates to \code{FALSE}, indicates that the increased complexity does 
not decrease generalization loss, and subsequently the boosting procedure is 
terminated.

The top-left panel of Figure \ref{fig:tree_model_fits} illustrates the fit of a model that converged after
$K=255$ iterations, and also shows a "convergence plot" (top-middle panel with test loss at different boosting iterations) that flattens out.
Indeed, repetitions of the same experiment with an increased number of training instances, $n$ (see top row of Figure \ref{fig:greedy_fun_fits}), shows that the test loss converges on average towards 1, indicated by the horizontal line.
This is the expected minimum value possible due to the standard Gaussian noise.
Furthermore, if considering the lower-right plot with the fit of the 200'th tree to 
the residuals at that iteration, we can see that the algorithm still finds 
information that is hard to see for the naked eye. The algorithm continues for another
55 iterations, that still manages to decrease test loss.

\subsection{The global-subset algorithm} \label{subsec:making space}
\begin{figure}[t!]
	\centering
	\includegraphics[width=1.0\textwidth,height=7cm]{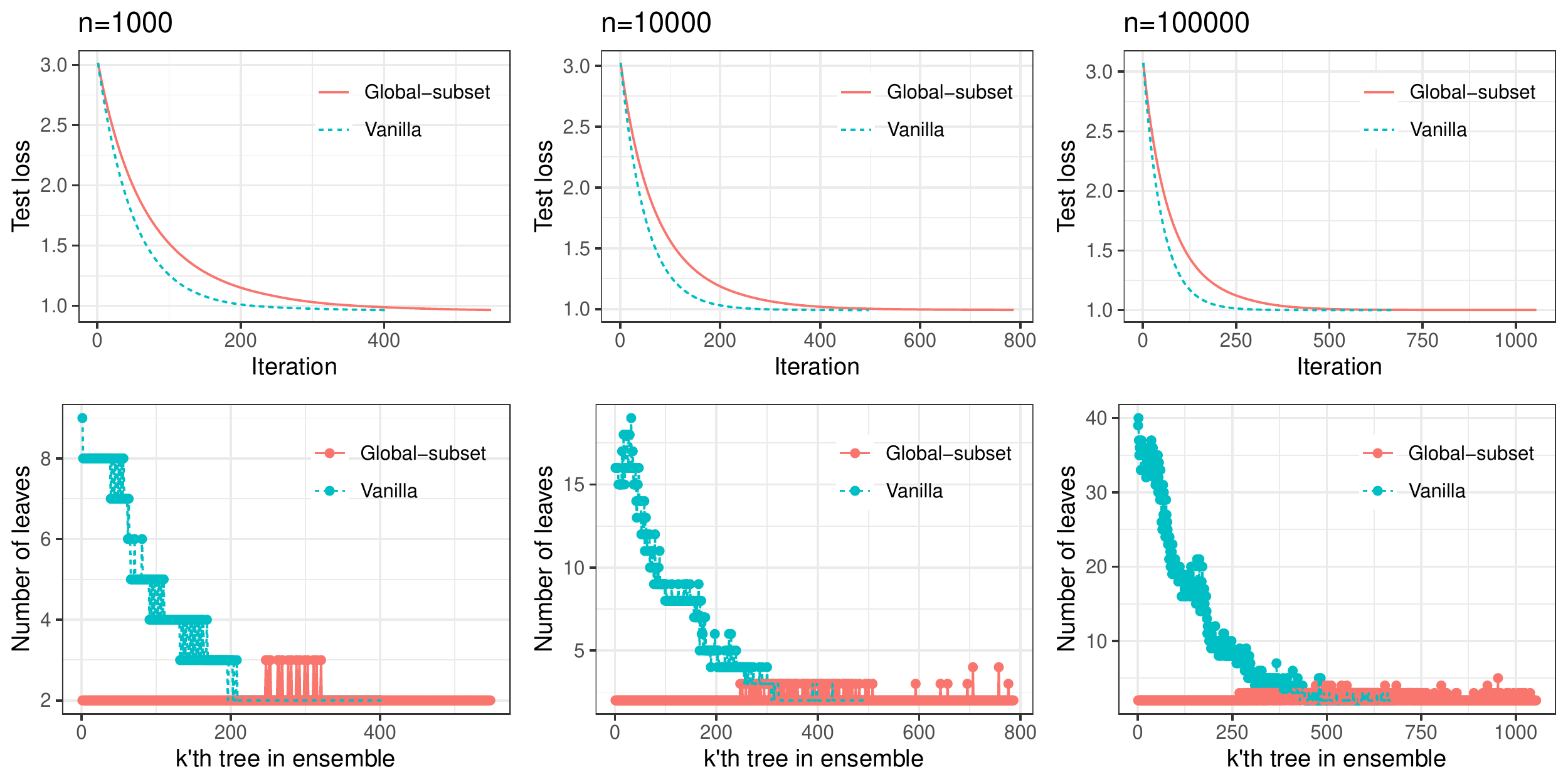}
	\caption{\label{fig:greedy_fun_fits} 
		Top: Test loss at different boosting iterations for the global-subset and vanilla
		algorithms training and testing with an increasing number of observations.
		Lower: The number of leaves in the $k$'th tree for the global-subset and vanilla 
		algorithms, also training and testing with the same number of observations as in
		the top row.
		The data generating process is the same as described in the caption of Figure \ref{fig:tree_model_fits}, but with the number of training and test observations 
		set to $n=\{1000, 10000, 100000\}$ for the three columns.
	}
\end{figure}

Equipped with the information criterion (\ref{eq:information-criteria-loss-reduction}), 
it is possible to construct a solution to the problem that each tree is optimized alone, mentioned as the third point in the introduction. 
For this subsection, denote the reduction in training loss from splitting some node $t$
at the $k$'th boosting iteration by ${}_{k}R_t$.
For example, the reduction from the root-node split at the same $k$'th iteration is denoted ${}_{k}R_1$ and the reduction from splitting the root-node at the 
$(k+1)$'th boosting iteration is denoted ${}_{k+1}R_1$.

The idea is rather simple, namely to compare the average generalization loss reduction
from a chosen split in the $k$'th boosting iteration, with that of the average generalization loss reduction we would obtain
from the root-split in the $(k+1)$'th boosting iteration if 
the aforementioned split was not performed and the recursive splitting at the $k$'th iteration terminated.
This then allows the tree-boosting algorithm to consider (in a greedy manner) all possible allowed changes
in function complexity of the ensemble, not just a deeper tree. 
The naive approach to do this -- at each possible split in the $k$'th iteration, 
temporarily terminate, and start on the $(k+1)$'th iteration for inspection of the 
root-split reduction in generalization loss -- is computationally infeasible.
Instead, notice that, as $\delta\to 0$, the 2'nd order gradient boosting approximation
to the loss is increasingly accurate, and  
that, at the limit $\delta\to 0$, we have $f_k=f_{k+1}$, as $f_k$ is 
scaled to zero by the learning rate.
Necessarily, we have ${}_{k}R_1 \approx {}_{k+1}R_1$ and $\tilde{C}_{{}_{k}R_1} \approx \tilde{C}_{{}_{k+1}R_1}$.

The quantities ${}_{k+1}R_1$ and $\tilde{C}_{{}_{k+1}R_1}$ may be used to adjust the right-hand-side 
of \eqref{eq:adaptive-tree-inequality} with the expected reduction in generalization loss of the 
next split, so that the recursive binary splitting terminates when splitting the next root-node is
more beneficial.
Using ${}_{k}R_1$ and $\tilde{C}_{{}_{k}R_1}$ as replacements, the reformulated split-stopping 
inequality yields
\begin{align}\label{eq:global-subset-inequality}
	\pi_t^{-1}
	\left({}_{k}R_t + \tilde{C}_{{}_{k}R_t}\right) 
	> \max\left\{0, {}_{k}R_1 + \tilde{C}_{{}_{k}R_1} \right\}.
\end{align}
The probability $\pi_t$ is introduced to adjust for the difference in the number of training observations, as the root works on the full dataset, while node $t$ necessarily works on some subset of the data. The inequality \eqref{eq:global-subset-inequality} is then employed as a replacement for- and in the exact same 
manner as \eqref{eq:adaptive-tree-inequality}

Figure \ref{fig:greedy_fun_fits} illustrates the practical difference in pathological learning behaviour: The data exhibits purely additive behaviour. 
\citet{friedman2000additive} argues for a model consisting only of tree-stumps in this case.
Both method converges to a test loss of approximately 1, the minimum expected test loss possible for a perfect model, for all values of $n$.
The difference lies in the complexity and number of trees.
The vanilla algorithm (using \eqref{eq:adaptive-tree-inequality}) builds each tree as if it was the last, and already at the first iteration, several regions of feature space will be split into sub-regions, seen from the plots in the lower row.
The global-subset algorithm, however, "looks ahead" and often evaluates that terminating the recursive binary splitting procedure and starting on a new boosting iteration is more beneficial.
Subsequently, trees are rarely complex and thus easier to interpret, but comes to the cost of 
a higher number of boosting iterations before terminating by the inequality \eqref{eq:automatic-early-stopping}.
This cost is decreased, however, as boosting iterations are overall faster than for the vanilla algorithm since individual trees are less complex.

\section{Using the agtboost package} \label{sec:using}
The goal of the \pkg{agtboost} package is to avoid expert opinions and computationally
costly brute force methods with regards to tuning the functional complexity
of GTB models. Usage should be as simple as possible.
As such, the package has only two main functions, \code{gbt.train} for
training an \pkg{agtboost} model, and a predict function that overloads the 
\code{predict} function in \proglang{R}.
The main responsibility of the user is to identify a "natural" loss-function and link-function.
To this end, \pkg{agtboost} also comes with a model validation function, \code{gbt.ksval}, which
performs a Kolmogorov-Smirnov test on supplied data, and a function for feature importance, \code{gbt.importance}, that functions similarly to ordinary feature importance functions (see for instance \citet{friedman2001elements}) but which calculates reduction in loss with respect to (approximate) generalization loss and not the ordinary training loss.
Due to implementation using \pkg{Rcpp modules}, saving and loading of \pkg{agtboost} cannot be done by the ordinary \code{save} and \code{load} functions in \proglang{R}, 
but is made possible through the functions \code{gbt.save} and \code{gbt.load}.
Table \ref{tab:loss_functions} gives an overview of the implemented loss functions 
in \pkg{agtboost}.
\begin{table}[t!]
	\centering
	\begin{tabular}{lllp{5.4cm}}
		\hline
		Type           & Distribution & Link   & Comment \\ \hline
		\code{mse}            & Gaussian      & $\mu=f(x)$       & Ordinary regression for continuous response \\
		\code{logloss}	& Bernoulli 	& $\log \left(\frac{\mu}{1-\mu}\right)=f(x)$ & Regression for classification problems \\
		\code{gamma::neginv}	& Gamma		& $-\frac{1}{\mu}=f(x)$	& Gamma regression for positive continuous response \\
		\code{gamma::log}	& Gamma		& $\log(\mu)=f(x)$	&  regression for positive continuous response \\
		\code{poisson}	& Poisson		& $\log(\mu)=f(x)$	& Poisson regression for count data exhibiting $Var(y|x)=E[y|x]$ \\
		\code{negbinom}	& Negative binomial		& $\log(\mu)=f(x)$	& For count data exhibiting overdispersion. \code{dispersion} must be supplied to \code{gbt.train} \\
		\hline
	\end{tabular}
	\caption{\label{tab:loss_functions} Overview of the loss functions available in 
		\pkg{agtboost}.}
\end{table}

Following is a walk-through of the \pkg{agtboost} package, applied to the \code{caravan.train} 
and \code{caravan.test} data \citep{van2000coil} that comes with the package and documented there.
The caravan dataset has a binary response, indicating purchase of caravan insurance, and 85 
socio-demographic covariates. Due to the nature of the response, classification using the 
\code{logloss} loss function is natural.
To train a GTB model, it is only needed to specify the \code{loss\_function} argument in \code{gbt.train}
\begin{CodeChunk}
\begin{CodeInput}
R> mod <- gbt.train(y = caravan.train$y, x = caravan.train$x,
		    loss_function = "logloss", verbose = 100)
\end{CodeInput}
\begin{CodeOutput}
it: 1  |  n-leaves: 3  |  tr loss: 0.2166  |  gen loss: 0.2167
it: 100  |  n-leaves: 2  |  tr loss: 0.1983  |  gen loss: 0.2019
it: 200  |  n-leaves: 3  |  tr loss: 0.1927  |  gen loss: 0.1987
it: 300  |  n-leaves: 2  |  tr loss: 0.1898  |  gen loss: 0.1978
\end{CodeOutput}
\end{CodeChunk}
Note the \code{verbose=100} argument, which creates output at the first and every 100'th
iteration. The output consists of the iteration number, the number of leaves of the $k$'th
tree, the training loss and approximate generalization loss.
By default, the global-subset algorithm using inequality \eqref{eq:global-subset-inequality} 
is used instead of \eqref{eq:adaptive-tree-inequality}, if the latter is preferred, specify
\code{algorithm = "vanilla"} as an argument to \code{gbt.train}.

\begin{figure}
	\centering
	\includegraphics[width=1.0\textwidth,height=7.5cm]{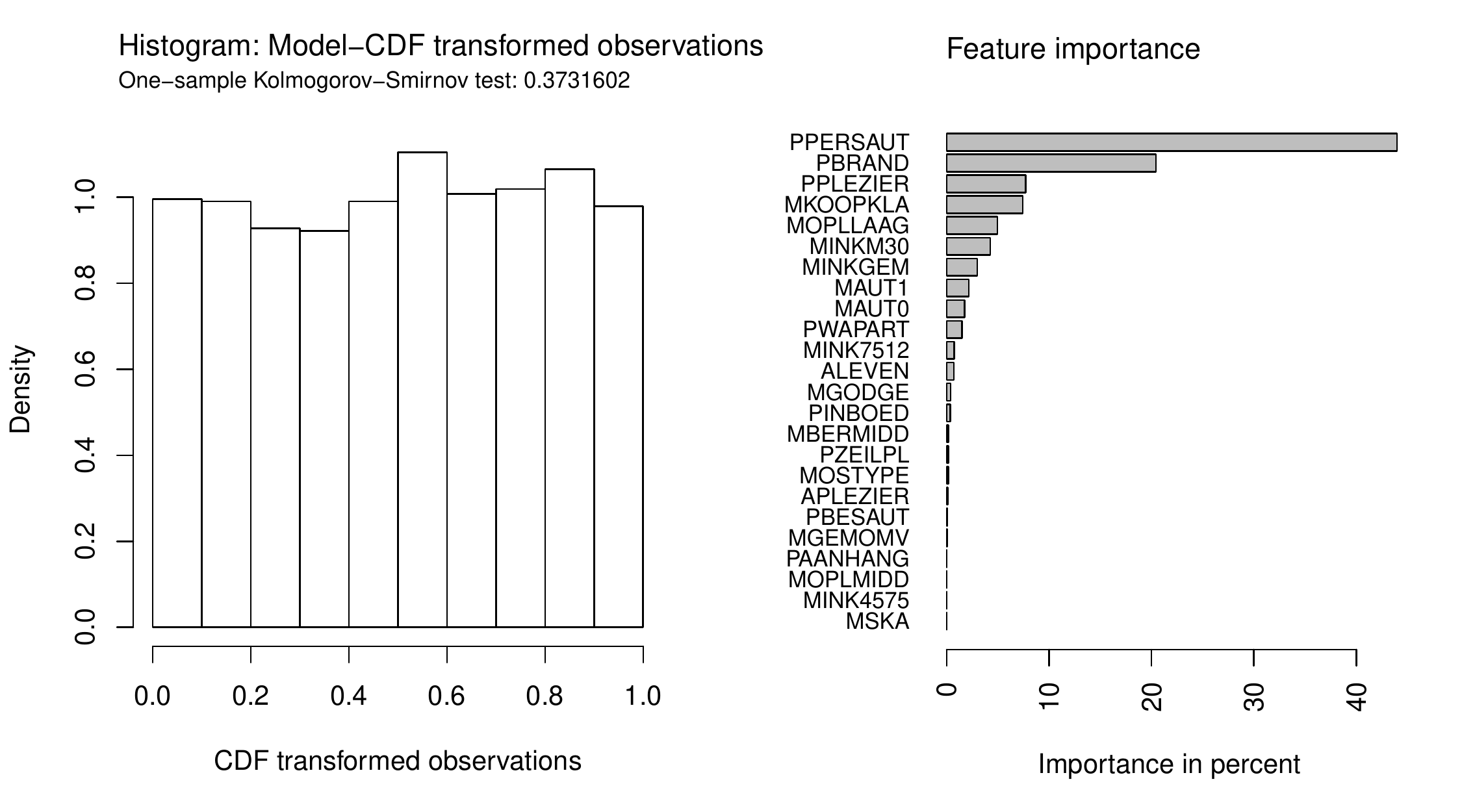}
	\caption{\label{fig:using_agtboost_caravan} 
		[left] Histogram generated by \code{gbt.ksval}, where test-response observations are transformed
		using \eqref{eq:discrete-uniform-transform}. As the histogram closely resembles the 
		histogram of standard uniformly distributed random variables, the model should not be discarded,
		something also indicated by the formal Kolmogorov-Smirnov test.
		[right] Feature importance plot generated by \code{gbt.importance}.
	}
\end{figure} 

The overloaded \code{predict} function may be used to check the fit on the 
training data, or to predict new data. 
To predict data, just pass the model object and the design matrix of data to be predicted
\begin{CodeChunk}
\begin{CodeInput}
R> prob_te <- predict(mod, caravan.test$x)
\end{CodeInput}
\end{CodeChunk}
It is often meaningful to do some formal goodness-of-fit test in addition to 
only visually inspecting the fit. A quite natural and general way to do this 
for loss functions associated negative log-likelihoods, is to use the Kolmogorov-Smirnov test \citep{kolmogorov1933sulla}.
The idea is to, for a continuous response, perform the CDF transform
\begin{align}\label{eq:continous-uniform-transform}
u_i = p\left(y\leq y_i;\hat{\theta}(x_i)\right),
\end{align}
preferably for test-data, and test the $u_i$'s against the standard uniform distribution, which holds if the model is correctly specified.
If the response is discrete, then employ the transform
\begin{align}\label{eq:discrete-uniform-transform}
	u_i = p\left(y \leq y_i-1; \hat{\theta}(x_i)\right) + Vp\left(y_i;\hat{\theta}(x_i)\right),
\end{align}
where $V\sim U(0,1)$, instead of the CDF transform.
All of this is implemented in the \code{gbt.ksval} function.
To apply it to the caravan data model, simply write
\begin{CodeChunk}
\begin{CodeInput}
R> gbt.ksval(object = mod, y = caravan.test$y, x = caravan.test$x)
\end{CodeInput}
\begin{CodeOutput}
Classification 
One-sample Kolmogorov-Smirnov test
data:  u
D = 0.021877, p-value = 0.3732
alternative hypothesis: two-sided
\end{CodeOutput}
\end{CodeChunk}
which produces output of the test-statistic, the p-value and the histogram in Figure \ref{fig:using_agtboost_caravan}.
Note that for multi-parameter distributions, such as the Gaussian, gamma and negative binomial,
the remaining parameters are assumed constant and maximum-likelihood estimates are produced during 
evaluation of the \code{gbt.ksval} function to allow for the transforms \eqref{eq:continous-uniform-transform}
and \eqref{eq:discrete-uniform-transform}.
The estimates will then be produced in the output.

In addition to the \code{gbt.ksval} function, the \code{gbt.importance} function can be used 
to inspect the model. This function produces a traditional feature importance plot (see e.g. \citet[Chap.~15.3.2]{friedman2001elements}), but different from other packages, the calculations are with respect to approximate generalization loss.
Formally, for non-leaf nodes $t$ in all trees in the ensemble, the value 
\begin{align*}
\delta (2-\delta) R_t + \delta \tilde{C}_{R_t}
\end{align*}
is added to the $j$-th element of a vector, where $j$ is the $j$-th feature used for the split.
The way in which \code{gbt.importance} calculates importance, and due to the sparse models produced
by \code{gbt.train}, know-how tricks such as inserting Gaussian noise-features are not necessary
when using \pkg{agtboost}.
The right plot in Figure \ref{fig:using_agtboost_caravan} produces the feature importance plot
for the caravan model. Note that only 24 of 85 possible features are used by the model.
Re-training the model with only these relevant features might improve the fit, as the 61 features not used 
by the model are noise that mask information and enlarges the absolute value of the 
information criterion \eqref{eq:information-criteria-loss-reduction}.

\section{Higgs big-data case study} \label{sec:case}

The two variants of \code{agtboost} is tested across increasing training sizes of a dataset, and 
their intrinsic behaviour with regards to reduction in loss, number of trees and leaves of trees, numbers of features used, and convergence across boosting iterations is studied.
We refer to models using inequality \eqref{eq:adaptive-tree-inequality} as "vanilla" models, and 
models using the global-subset algorithm \eqref{eq:global-subset-inequality} as "global-subset" models.
To this end, the Higgs dataset\footnote{https://archive.ics.uci.edu/ml/datasets/HIGGS} is used.
The Higgs data consists of 11 million observations of a binary response and 28 continuous features.
The first 10 million observations are used for training, and the last million for testing for which
results are reported. The training set is sampled randomly without replacement for $n=\{10^{2},10^{3},10^{4},10^{5}\}$ observations, and trained on the respective
training indices. Tests and reported results are always done on the one-million sized test-set.

\begin{table}[t!]
\centering
\begin{tabular}{ccccccc}
\hline
Algorithm & Loss & AUC  & Time & \#trees & \#leaves & \#features \\ 
\multicolumn{7}{c}{$n=100$}\\ 
vanilla & 0.6728 & 0.5942 & 0.1417 & 32 & 64 & 1\\ 
global-subset & 0.6734 & 0.5942 & 0.1386 & 30 & 60 & 1\\ 
\hline 
\multicolumn{7}{c}{$n=1000$}\\ 
vanilla & 0.6483 & 0.703 & 2.335 & 162 & 422 & 7\\ 
global-subset & 0.6437 & 0.7071 & 1.623 & 184 & 504 & 8\\ 
\hline 
\multicolumn{7}{c}{$n=10000$}\\ 
vanilla & 0.5692 & 0.7796 & 1.191 & 684 & 4976 & 22\\ 
global-subset & 0.5708 & 0.7781 & 1.201 & 768 & 4008 & 18\\ 
\hline 
\multicolumn{7}{c}{$n=100000$}\\ 
vanilla & 0.5317 & 0.8087 & 32.57 & 1055 & 43908 & 28\\ 
global-subset & 0.5321 & 0.8085 & 34.17 & 1176 & 29105 & 28\\ 
\hline 
\end{tabular}
\caption{\label{tab:higgs} 
Results of the two algorithms available in \pkg{agtboost} for different 
number of training samples available. 
The Loss (Logloss) and AUC is computed on the test set with 1 million observations.
The remaining columns are the total number of trees in the models, the total
number of leaves summed up over all trees, and the total number of features
used by the models.
Loss metrics and computation times are almost identical, while complexity and construction
differs.
}
\end{table}

\begin{figure}[t!]
\centering
\includegraphics[width=1.0\textwidth,height=6cm]{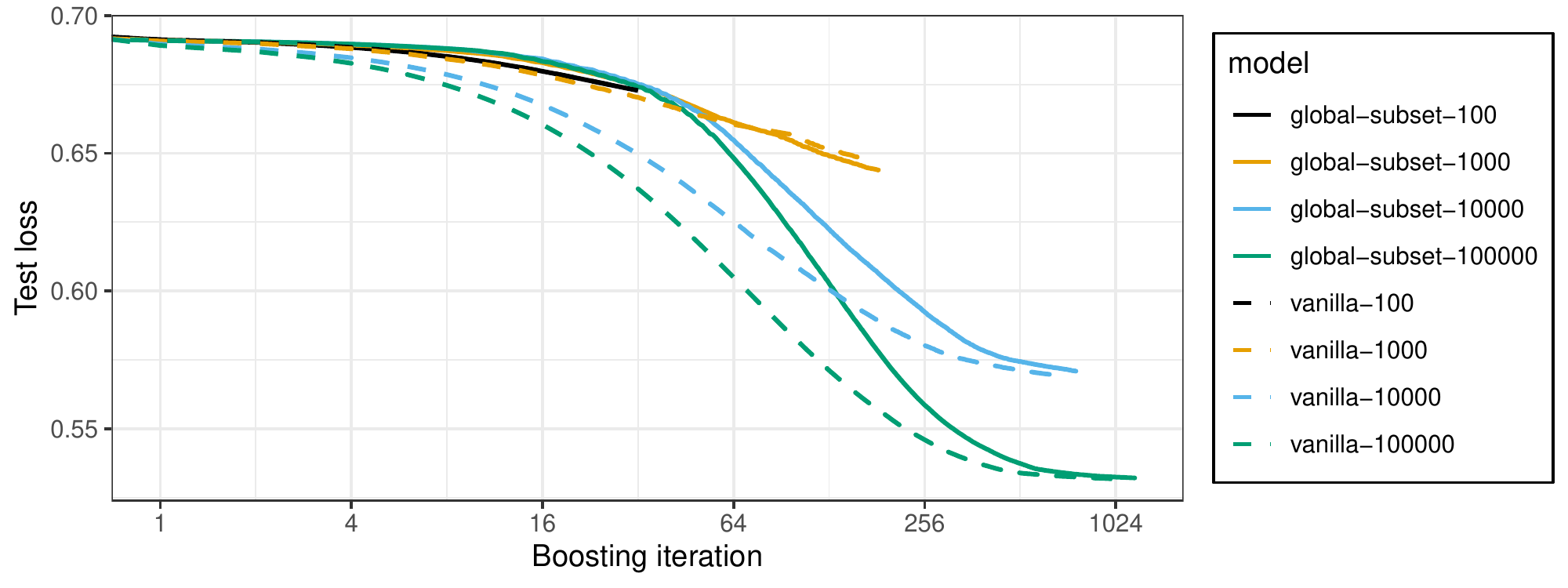}
\caption{\label{fig:model-convergence} 
Test loss as a function of boosting iterations, for the vanilla and global-subset
models trained on $n\in\{100, 1000, 10000, 100000\}$ training observations.
More training observations imply more information, which allow for lower points of convergence
and a greater number of boosting iterations.
The methods stop at different boosting iterations.
Notice the log scale on the horizontal axis. 
}
\end{figure}

\begin{figure}[t!]
\centering
\includegraphics[width=1.0\textwidth,height=6cm]{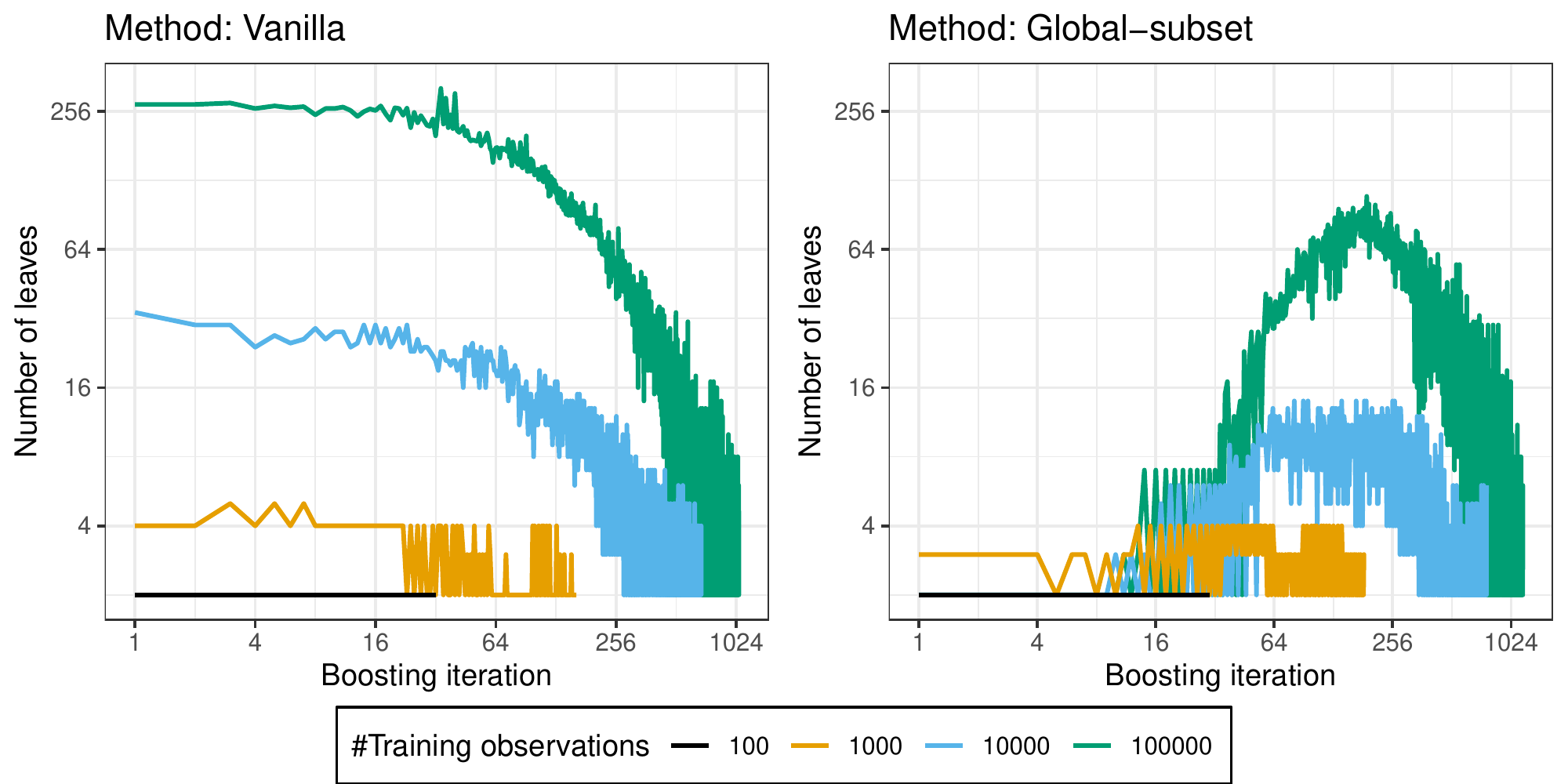}
\caption{\label{fig:model-nleaves} 
The number of leaves for the $k$'th tree (boosting iteration $k$) for different 
\pkg{agtboost} models trained on the Higgs data with a varying number of training observations.
Notice the log scale on both the vertical and horizontal axes. 
}
\end{figure}


In the "Loss" column of Table \ref{tab:higgs} it is seen that the test-loss is decreasing in the size of 
the training set, as it should. Figure \ref{fig:model-convergence} compliments this result:
Each model is seen to converge and values of test loss flatten out as a function of boosting iterations.
But, as the training set increases,
more information in the data is present that allow lower points of convergence. None of the models 
are seen to overfit in the number of boosting iterations, as the curves never increase.

While the two variants of \pkg{agtboost} converge to similar results in terms of test-loss,
and the methods take a similar amount of time (column 5 of Table \ref{tab:higgs}),
their behaviour during training and the complexity of the resulting models differ.
Figure \ref{fig:model-nleaves} shows two different ways of learning the structural signal in the data.
The early trees of the vanilla algorithm start with deep trees, and as the signal is learned, trees become smaller.
Trees from the global-subset algorithm, on the other hand, start out with mere tree stumps, and then increase in size as interaction effects become more beneficial to learn than additive relationships.
Trees do however not reach the depth of the deepest early trees of the vanilla algorithm.
As interaction effects are taken into the model, these trees also become smaller before convergence of the boosting algorithm.
The total number of trees and leaves of the models are shown in columns 6 and 7 in Table \ref{tab:higgs}.
While the global-subset algorithm produces models with a larger number of trees than the vanilla algorithm, the total number of leaves is typically smaller, but without a loss of accuracy.
As sparsity is a good defence against the curse of dimensionality, the performance of the global-subset
algorithm might become more evident on big-$p$ small-$n$ datasets.

\section{Discussion} \label{sec:discussion}

This paper describes \pkg{agtboost}, an \proglang{R} package for gradient tree boosting solving
regression-type problems in an automated manner.
The package takes advantage of recent innovations in information theory with regards to the splits
in gradient boosted trees \citet{lunde2020information}, implements these in \proglang{C++} for fast computation
and employs \pkg{RcppEigen} for bindings to \proglang{R} which provides user-friendly application.
The package comes with two different utilizations of the information criterion \eqref{eq:information-criteria-loss-reduction}, that vary little in final accuracy and in training time but vary in terms for individual tree-size and complexity.
The package can be used for early exploratory data analysis for selecting features and an appropriate loss-function, but also for building a final highly predictive model.


\bibliography{refs}


\newpage

\begin{appendix}

\end{appendix}


\end{document}